\documentclass[conference]{IEEEtran}
\IEEEoverridecommandlockouts
\usepackage{cite}
\usepackage{amsmath,amssymb,amsfonts}
\usepackage{algorithmic}
\usepackage{graphicx}
\usepackage{textcomp}
\usepackage{authblk}
\usepackage{gensymb}

\usepackage{hyperref}
\usepackage{scrextend}
\usepackage{changepage}
\usepackage[margin=0.75in]{geometry}
\usepackage{multirow}
\usepackage{amsfonts}
\usepackage[table]{xcolor}
\usepackage{booktabs}

\hypersetup{
     colorlinks   = true,
     linkcolor    = blue,
     citecolor    = black
}

\def\BibTeX{{\rm B\kern-.05em{\sc i\kern-.025em b}\kern-.08em
    T\kern-.1667em\lower.7ex\hbox{E}\kern-.125emX}}

\makeatletter
\newsavebox\myboxA
\newsavebox\myboxB
\newlength\mylenA 
\newcommand*\xoverline[2][0.75]{%
    \sbox{\myboxA}{$\m@th#2$}%
    \setbox\myboxB\null% Phantom box
    \ht\myboxB=\ht\myboxA%
    \dp\myboxB=\dp\myboxA%
    \wd\myboxB=#1\wd\myboxA% Scale phantom
    \sbox\myboxB{$\m@th\overline{\copy\myboxB}$}%  Overlined phantom
    \setlength\mylenA{\the\wd\myboxA}%   calc width diff
    \addtolength\mylenA{-\the\wd\myboxB}%
    \ifdim\wd\myboxB<\wd\myboxA%
       \rlap{\hskip 0.5\mylenA\usebox\myboxB}{\usebox\myboxA}%
    \else
        \hskip -0.5\mylenA\rlap{\usebox\myboxA}{\hskip 0.5\mylenA\usebox\myboxB}%
    \fi}
\makeatother

\begin{document}
\title{\vspace{0.25in}{\huge \textbf{Is That a Chair? Imagining Affordances Using Simulations of an Articulated Human Body}
}}

\author{Hongtao Wu,~\IEEEmembership{Student Member,~IEEE}, Deven Misra, and Gregory S. Chirikjian,~\IEEEmembership{Fellow,~IEEE}
\thanks{H. Wu is with the Laboratory for Computational Sensing and Robotics (LCSR), Johns Hopkins University, Baltimore, MD 21218, USA. {\tt \{hwu67\}@jhu.edu} D. Misra is with the Physics Department, Reed College, Portland, OR 97202, USA. G. S. Chirikjian is with the Department of Mechanical Engineering, National University of Singapore, Singapore and LCSR, Johns Hopkins University, Baltimore, MD 21218, USA. G. S. Chirikjian is the corresponding author. {\tt \{mpegre\}@nus.edu.sg}
}}

\maketitle

\begin{abstract} 
For robots to exhibit a high level of intelligence in the real world, they must be able to assess objects for which they have no prior knowledge. Therefore, it is crucial for robots to perceive object affordances by reasoning about physical interactions with the object. In this paper, we propose a novel method to provide robots with an ability to imagine object affordances using physical simulations. The class of chair is chosen here as an initial category of objects to illustrate a more general paradigm. In our method, the robot ``imagines" the affordance of an arbitrarily oriented object as a chair by simulating a physical sitting interaction between an articulated human body and the object. This object affordance reasoning is used as a cue for object classification (chair vs non-chair). Moreover, if an object is classified as a chair, the affordance reasoning can also predict the upright pose of the object which allows the sitting interaction to take place. We call this type of poses the \textit{functional pose}. We demonstrate our method in chair classification on synthetic 3D CAD models. Although our method uses only 30 models for training, it outperforms appearance-based deep learning methods, which require a large amount of training data, when the upright orientation is not assumed to be known \textit{a priori}. In addition, we showcase that the functional pose predictions of our method align well with human judgments on both synthetic models and real objects scanned by a depth camera.
\end{abstract}

\section{Introduction}

Object affordances play an essential role in object perception \cite{gibson2014ecological}. Experiments show that object affordances are more compelling in infants' conceptual development of object perception than colors, textures, and other perceptual cues \cite{oakes2008function, nelson1973some}. Gibson \cite{gibson2014ecological} contends that predicting object affordances is more important than predicting object class labels. This is also true from a robotic perspective. For instance, when a humanoid robot is confronted with an object classified as a chair, the label of  ``chair" would not provide the robot with the knowledge of how to sit on it; while knowing if it is ``sittable" and the pose which affords the sitting function offer more informative cues for object perception and interaction.

\begin{figure}[!htp]
\centering
\includegraphics[scale = 0.21]{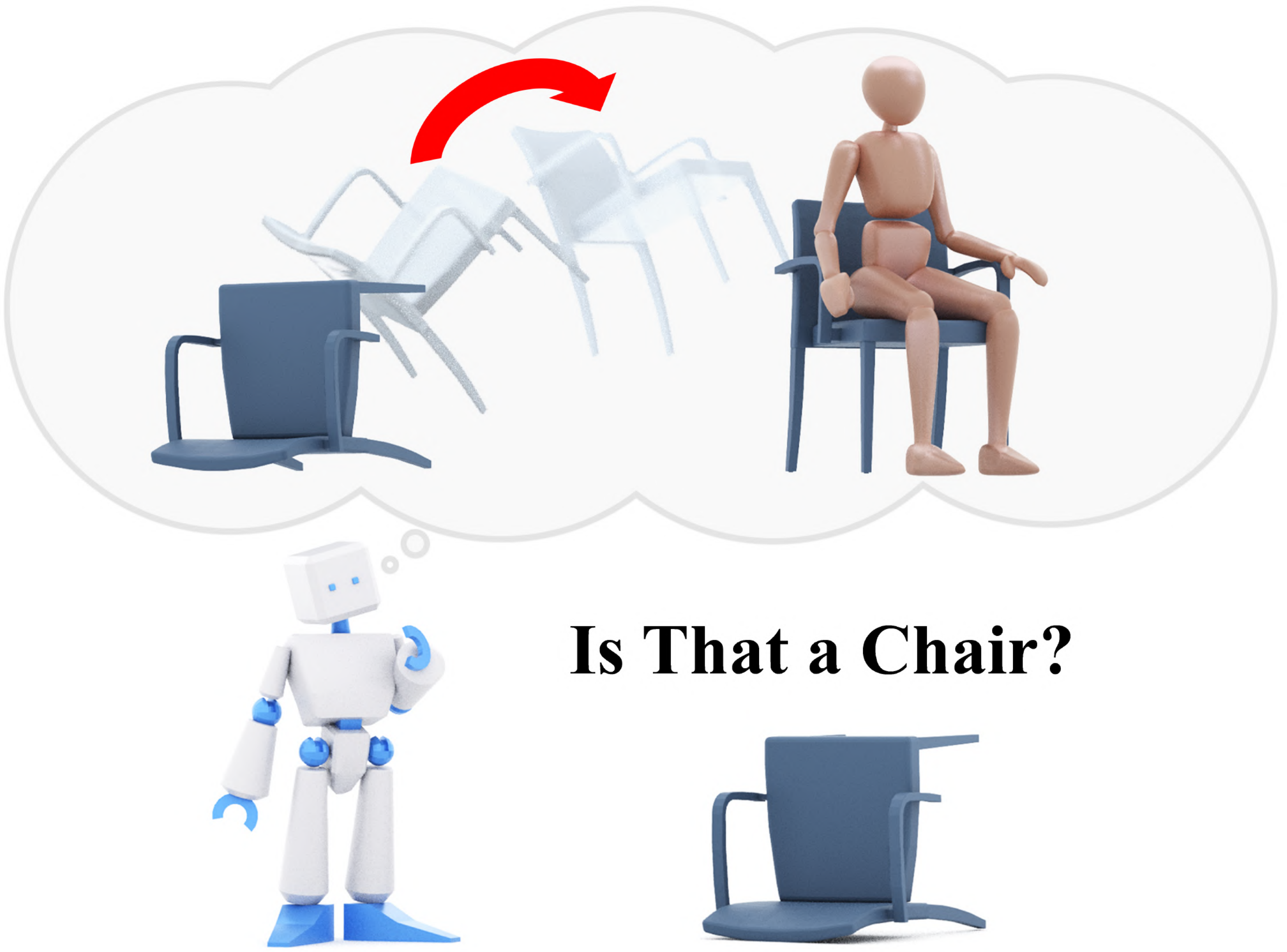}
\caption{Chair Affordance Imagination. Our method \textit{imagines} the object affordance as a chair by physically simulating a human agent sitting on it. The agent is simplified as an articulated human body. The agent and the object are subject to constraints of geometry and physics (\textit{e.g.}, gravity, friction, collision, and inertia). More details can be found on our project page: \url{https://chirikjianlab.github.io/chairimagination/}}
\label{figure1}
\vspace{-0.2cm}
\end{figure}

\begin{figure*}[!htp]
\centering
\includegraphics[scale = 0.545]{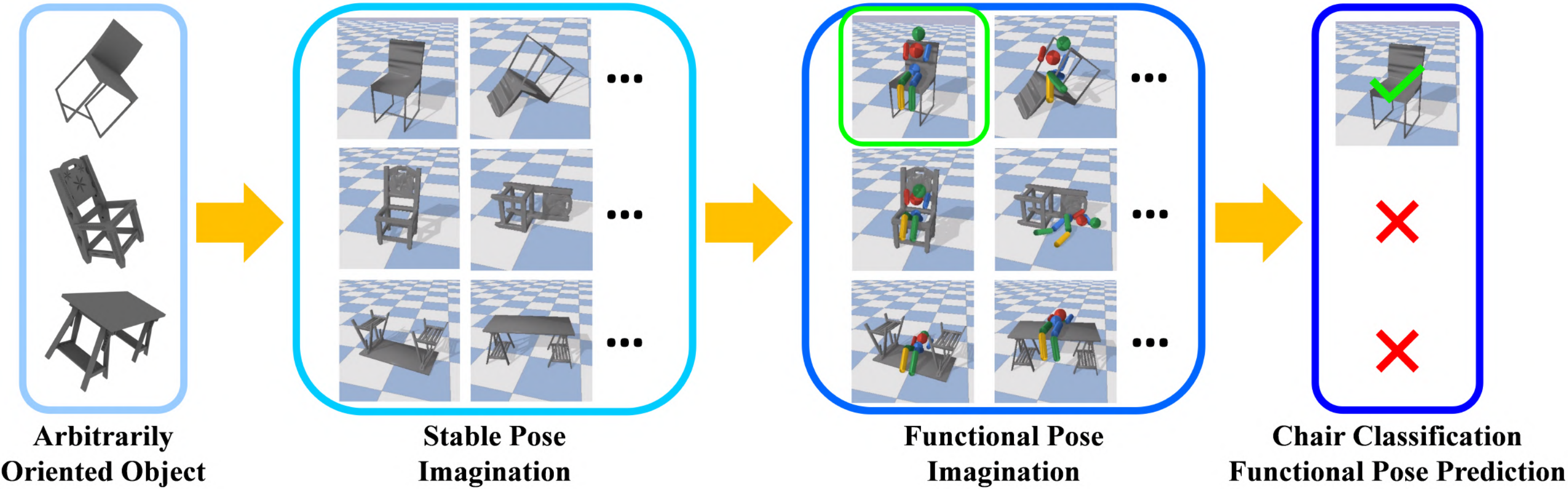}
\caption{Imagination Pipeline. The imagination consists of two steps: stable pose imagination and functional pose imagination. Given an arbitrarily oriented object, we first imagine its stable poses. Then, we proceed to imagine the functional poses by performing a sitting interaction between a human agent and the object in the stable poses found in the first step. The \textit{sitting quality} for each stable pose is reasoned to check if there exists any functional pose which affords sitting. The three objects shown in the figure include a chair, a broken chair without a seat, and a table. Only the chair has functional poses (the green box in the Functional Pose Imagination) which afford sitting.}
\label{figure2}
\vspace{-0.2cm}
\end{figure*}

Despite the advantages of affordance-based object perception, the majority of work on object perception is appearance-based. These methods can be limiting when considering object classes with a large intra-class appearance variation such as chairs. Moreover, in the case where an object appears similar to a class but fails to afford the most salient functionality of the class, the appearance-based method would be challenged. For example, consider a ``broken chair" without a seat (Figure \ref{figure2}) or a ``toy chair" for dollhouse play.

 Our primary contribution is a novel method which \textit{imagines} an object's affordance as a chair by physically simulating a human agent interacting with the object (Figure \ref{figure1}). Our method is inspired by the fact that human brains perform mental simulations to reason high-level physical interactions of complex systems \cite{hegarty2004mechanical, gentner2014mental, battaglia2013simulation}. We want to endow robots with an analogous ``mental" capability which \textit{imagines} possible physical interactions of a class and deduces an object's affordance as the class accordingly. Therefore, we first define the physical interaction to imagine for the class of chair.

The Merriam-Webster\footnote{https://www.merriam-webster.com/dictionary/chair} dictionary defines a chair as: ``a seat typically having four legs and a back for one person". However, this human-centric definition does not provide any affordance information or guidance on how the object can be interacted with. To be more robot-centric, we propose an \textit{interaction-based definition} of a chair: 
\begin{adjustwidth}{1cm}{1cm}
``an object which can be stably placed on a flat horizontal surface in such a way that a typical human is able to \textit{sit}\footnote{https://www.collinsdictionary.com/dictionary/english/sit} ($\rightarrow$ to adopt or rest in a posture in which the body is supported on the buttocks and thighs and the torso is more or less upright) stably above the ground."

\end{adjustwidth}
This definition provides affordance information on how to interact with an object, \textit{i.e.}, if the object is a chair, one can sit on it. Moreover, if the object is a chair, it is able to afford sitting on a flat surface in some poses which are stable. We define this type of poses as the \textit{functional pose}. These poses are associated with the designed upright orientation of the object. Figure \ref{figure2} shows our imagination pipeline. Using this imagination, we are able to quantify the sitting affordance of an object. This quantification can be further used as a cue for chair vs non-chair classification (hereafter referred to as chair classification). In addition, the functional pose which affords sitting can also be predicted if the object is categorized as a chair. 
We demonstrate our method's performance on the chair classification and functional pose prediction tasks. For chair classification, we compare with two state-of-the-art appearance-based deep learning methods. Results show that our method outperforms the state of the art on synthetic 3D model data when the upright orientation of the model is not assumed. For functional pose prediction, we show that our method's predictions align well with human judgments on both synthetic models and real objects scanned by a consumer-grade depth camera.

\section{Related Work}

\textbf{Affordance-based Object Classification.}
There is a growing interest in classifying objects with object affordances \cite{ho1987representing, yu2015fill, hinkle2013predicting, grabner2011makes, bar2006functional, seib2016detecting, sawatzky2017weakly, roy2016multi, zhu2014reasoning}. Hinkle and Olson \cite{hinkle2013predicting} simulate dropping spheres onto objects and classify objects based on the final configuration of the spheres using the support vector machine (SVM). This differs from our approach which introduces a simulated human agent for exploring human-centric affordances. Grabner \textit{et al.} \cite{grabner2011makes} fit a human mesh model onto an upright chair and estimates the vertex nearest distance and triangle intersection between them to measure the sitting affordance. In \cite{bar2006functional} and \cite{seib2016detecting}, the sitting affordance of an object is detected by searching an embodied agent's configuration space. Only collision detection is considered in the search.
Unlike \cite{grabner2011makes, seib2016detecting, bar2006functional} which are solely based on geometry, our method considers not only geometric constraints but also physical properties (mass, inertia matrix, restitution, friction, \textit{etc.}) of the agent and the interacting object. In addition, we predict the \textit{functional pose} of a chair from an arbitrary orientation while the above methods all assume upright orientation in classfication. The functional pose prediction can benefit the affordance reasoning in the case where the chair is not in the pose to afford the functionality of sitting.

\textbf{Learning-based Affordance Detection.}
Learning-based methods have been widely applied to detect object affordances \cite{nguyen2016detecting, do2018affordancenet, aldoma2012supervised, myers2015affordance, desai2013predicting, tenorth2013decomposing, fu2008upright, manuelli2019kpam, manuelli2019kpam, mar2017can}. Aldoma \textit{et al.} \cite{aldoma2012supervised} detect the ``0-order affordance" of an object in a supervised learning manner. \cite{myers2015affordance, desai2013predicting, tenorth2013decomposing} focus on detecting functional regions and parts of objects by learning geometric features.  Mar \textit{et al.} \cite{mar2017can} use a self-supervised learning method to learn tool affordances based on the 3D geometry with Self-Organizing Maps (SOMs). Manuelli \textit{et al.} \cite{manuelli2019kpam} recently propose keypoint affordances to reinforce purposeful robot manipulation. Instead of relying on learning, our work encodes object affordances by simulating physical interactions with an object to measure its potential to afford a functionality.

\textbf{Physical Reasoning and Scene Understanding.}
Object affordances have also been used to understand physics in real world scenes \cite{zhao2013scene, zhu2016inferring, zhu2015understanding, jia20133d, zheng2013beyond, battaglia2013simulation, ruiz2018can, wu2015galileo, ruiz2018what}. 
% \cite{jia20133d, zheng2013beyond, zheng2014detecting} reason about the physical stability of objects in the scene. 
Battaglia \textit{et al.} \cite{battaglia2013simulation} propose an ``intuitive physics engine" to simulate physics in natural scenes and explain human mental models for understanding the real world. \cite{ruiz2018can} and \cite{ruiz2018what} leverage deep 3D saliency and interactive tensors to detect object afforadnces in 3D scene.  Our work diverges from these approaches by using object affordances for object classification and functional pose prediction instead of physical reasoning or scene understanding.

\textbf{Affordance-based Shape Analysis.}
Object affordances have also been explored with geometry-based shape analysis\cite{pirk2017understanding, hu2016learning, kim2014shape2pose, hu2015interaction}.
Pirk \textit{et al.} \cite{pirk2017understanding} capture the interaction between a motion driver and a static object by obtaining the ``interaction landscape" with animations. Hu \textit{et al.} \cite{hu2016learning} seek to understand local object affordances by learning the correlation between local geometric properties and object functionalities. Instead of geometric analysis, our method uses physics to deduce object affordances.

\section{Method}

Given an object in arbitrary orientation, our goal is to find if there exists any \textit{functional pose} which affords the sitting function. 
We attach the body frame to the center of mass of the object and align the axes parallel to those of the world frame. All objects are considered as rigid bodies. A rigid body transformation can be specified by $g=(R, \textbf{p}) \in SE(3)$. $R \in SO(3)$ is a rotation matrix which can be parameterized using the Euler angles: $R = R(\alpha, \beta, \gamma)$. We use the x-y-z extrinsic rotation convention for Euler angles throughout this paper. $\alpha, \beta, \gamma$ correspond to the roll, pitch, and yaw of the object.  $\textbf{p} \in \mathbb{R}^3$ is the translation vector which can be specified by the body frame's three coordinates in the world frame $\textbf{p} = [x, y, z]^T$.

A rigid body has infinitely many poses in $SE(3)$. It is computationally costly to finely search the whole $SE(3)$ to find functional poses. The \textit{interaction-based definition} also indicates that the functional pose is necessarily stable. Therefore, we first find a set of stable poses $G_{s} \subset SE(3)$ of the object. We then perform the sitting interaction on each stable pose $g_s \in G_{s}$ to find the functional pose $g_f \in G_{s}$.

\subsection{Stable Pose Imagination}
\label{sec:stable_pose}
\begin{figure}[!htp]
\centering
\includegraphics[scale = 0.34]{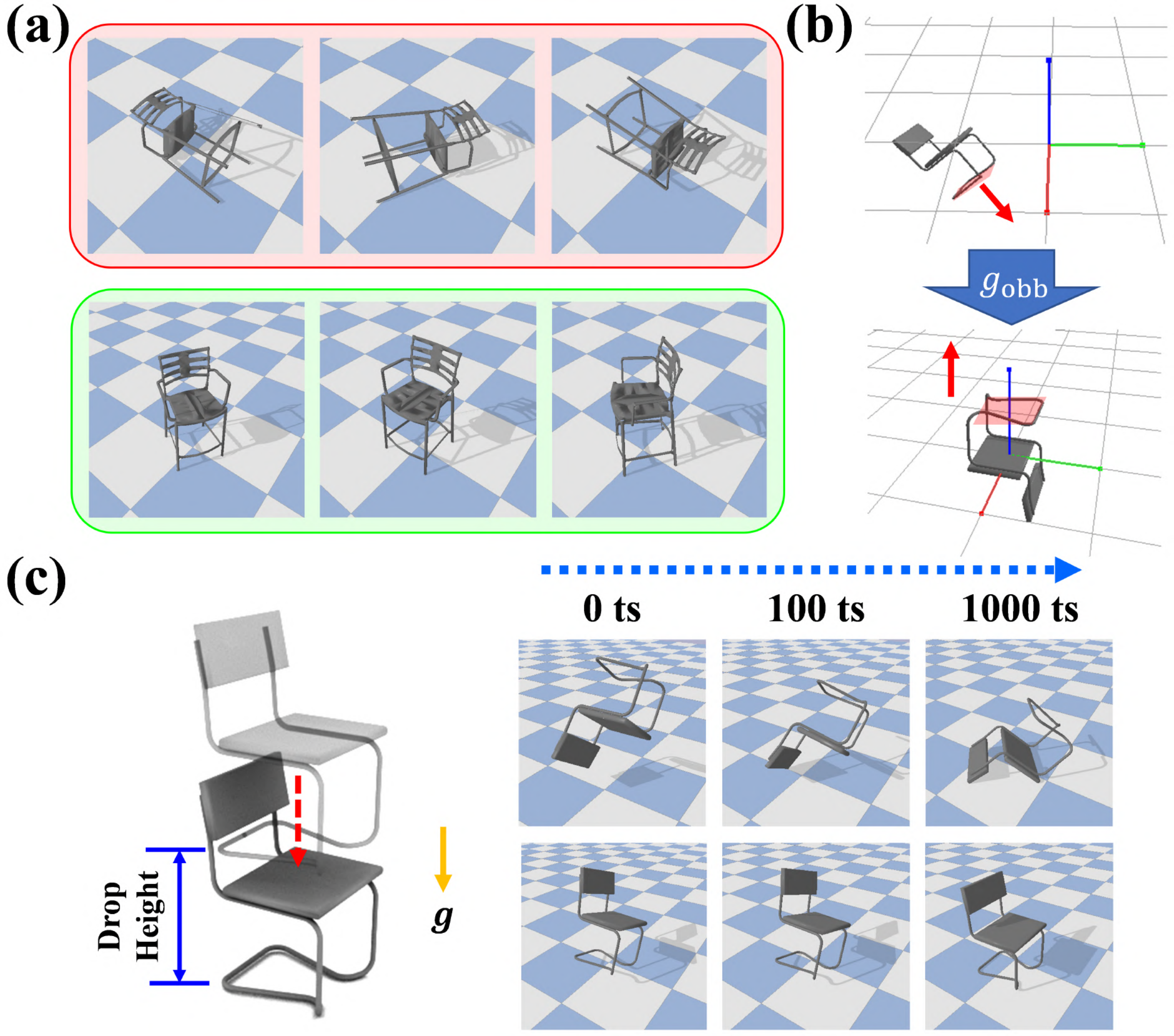}
\caption{Stable Pose Imagination. (a) A tipped over chair (red box) would not afford sitting regardless of its $\gamma$, $x$, and $y$. An upright chair (green box) can always afford sitting regardless of its $\gamma$, $x$, and $y$. (b) The upper and lower images show a chair object before and after applying $g_{\textrm{obb}}$, respectively. OBB is the minimum bounding box of which one of its surfaces is coincident with one of the surfaces of the object's convex hull. The red plane and the red arrow indicate the surface in contact with the ground when the chair is in the \textit{functional pose}. $g_{\textrm{obb}}$ transforms this surface to be normal to the z-axis in this case. (c) For each enumerated orientation, the object is dropped from a height equal to half the length of the diagonal of the OBB plus 5cm. The right figure shows the snapshots of the dropping simulation at the 0th, 100th, and 1000th time step, respectively.}
\label{figure3}
\vspace{-0.2cm}
\end{figure}

We simulate dropping an object with different initial orientations on a flat plane to find the stable poses of the object. However, there can be infinitely many such poses. We notice that many of these poses are functionally equivalent. For example, all translations in the x-y plane and all rotations around the z-axis of the world frame cause no change in the functionality of a chair pose, as long as the human agent positions and orients its stance in $SE(2)$ correctly. Figure \ref{figure3}(a) shows two examples. Therefore, we say two stable poses are \textit{equivalently stable} if their roll $\alpha$, pitch $\beta$, and z-axis coordinate $z$ in the world frame are equal. In the simulation, we consider two poses to be \textit{equivalently stable} if:
\begin{equation}\label{eq:1}
\|R(\alpha, \beta, 0)-R(\alpha', \beta', 0)\| < \Delta R_{\textrm{thr}}^{\textrm{es}}    
\end{equation}
\begin{equation}\label{eq:2}
|z - z'| < \Delta z_{\textrm{thr}}^{\textrm{es}}  
\end{equation}
where $\|A\|$ is the Frobenius norm of the matrix $A$; $R(\alpha, \beta, 0)$ and $R(\alpha', \beta', 0)$ are the rotation matrices which represent the roll and pitch of the two poses given their Euler angles $(\alpha, \beta, \gamma)$ and $(\alpha', \beta', \gamma')$; $z$ and $z'$ are the z-axis coordinates of the two poses. 
$\Delta R_{\textrm{thr}}^{\textrm{es}}, \Delta z_{\textrm{thr}}^{\textrm{es}} \in \mathbb{R}$ are two thresholds.

\textbf{Object Bounding Box Transformation.}
Given an arbitrarily oriented object, we first compute the minimum volume oriented bounding box (OBB) and apply the rigid body transformation $g_{\textrm{obb}}$ which aligns the center of the OBB with the origin of the world frame and rotates the object such that the edges of the OBB are parallel to the coordinate axes of the world frame (Figure \ref{figure3}(b)). We apply $g_{\textrm{obb}}$ because we notice that when chairs are in a functional pose, the surface in contact with the ground is heuristically coincident with one of the surfaces of its OBB. $g_{\textrm{obb}}$ allows the enumeration of orientations which are very close to the orientation of the functional pose and thus would stabilize to a functional pose after being dropped.

\textbf{Dropping Simulation.}
After the OBB transformation, we enumerate the orientation of the object by varying its roll $\alpha$ and pitch $\beta$ in a discrete increment $\Delta \alpha = \Delta \beta = \frac{\pi}{10}$ while keeping $\gamma=0$. Figure \ref{figure3}(c) shows the simulation setup. The simulation duration for each drop is 1000 time steps (1/240 seconds per time step). The object's pose in the last time step is considered a stable pose if in the last 50 time steps:
\begin{equation}\label{eq:3}
    \sum_{i} \|R_{i} - R_{i-1}\| \leq \Delta R_{\textrm{thr}}^{\textrm{s}}
\end{equation}
\begin{equation}
    \sum_{i} \|\textbf{p}_{i} - \textbf{p}_{i-1}\| \leq \Delta p_{\textrm{thr}}^{\textrm{s}}
\end{equation}
where $R_i$ and $\textbf{p}_{i}$ are the rotation matrix and position of the object in the i-th time step, respectively; $ \Delta R_{\textrm{thr}}^{s},  \Delta p_{\textrm{thr}}^{s} \in \mathbb{R}$ are two thresholds.
We initialize $G_s$ as an empty set. If a newly found stable pose is equivalently stable to a pose already in $G_s$, we discard it. Otherwise, we append it to the set.

\subsection{Functional Pose Imagination}
Sitting is performed on each stable pose $g_s=(R_s, \textbf{p}_s)$ to find the functional pose. We denote $R_s = R(\alpha_s, \beta_s, \gamma_s)$ and $\textbf{p}_s=[x_s, y_s, z_s]^T$. Figure \ref{figure4} shows our sitting simulation setting. We simplify the human agent as an articulated human body. We trim off the arms and feet because they are not substantial in defining a sitting configuration \cite{forssberg1994postural, kerr1997analysis}. We set appropriate limits, friction, and damping for each joint to avoid configurations which are not physiologically capable for a typical human \cite{roaas1982normal}.

For each $g_s$, the object's orientation is enumerated by fixing $\alpha = \alpha_s$, $\beta = \beta_s$ while varying $\gamma$ from $[0, 2\pi)$ in a discrete increment $\Delta \gamma = \pi/9$. The agent's pelvis is placed on a horizontal plane 15cm above the current axis-aligned bounding box (AABB) of the object. For each enumerated orientation, we sample three positions on the plane (the origin and two positions with a translation of $L_{\textrm{sit}}$ and $2L_{\textrm{sit}}$ along the x-axis, respectively) and freely drop the agent onto the object from each position. We regard each drop as a \textit{sitting trial}. In total, we conduct 54 sitting trials for each $g_s$. Since the sitting affordance depends on the agent's size and where the agent sits \cite{gibson2014ecological}, we scale the agent's size and $L_{\textrm{sit}}$ linearly with respect to the size of the object's OBB. 

\subsection{Sitting Affordance Model}
\begin{figure}[!htp]
\centering
\includegraphics[scale = 0.28]{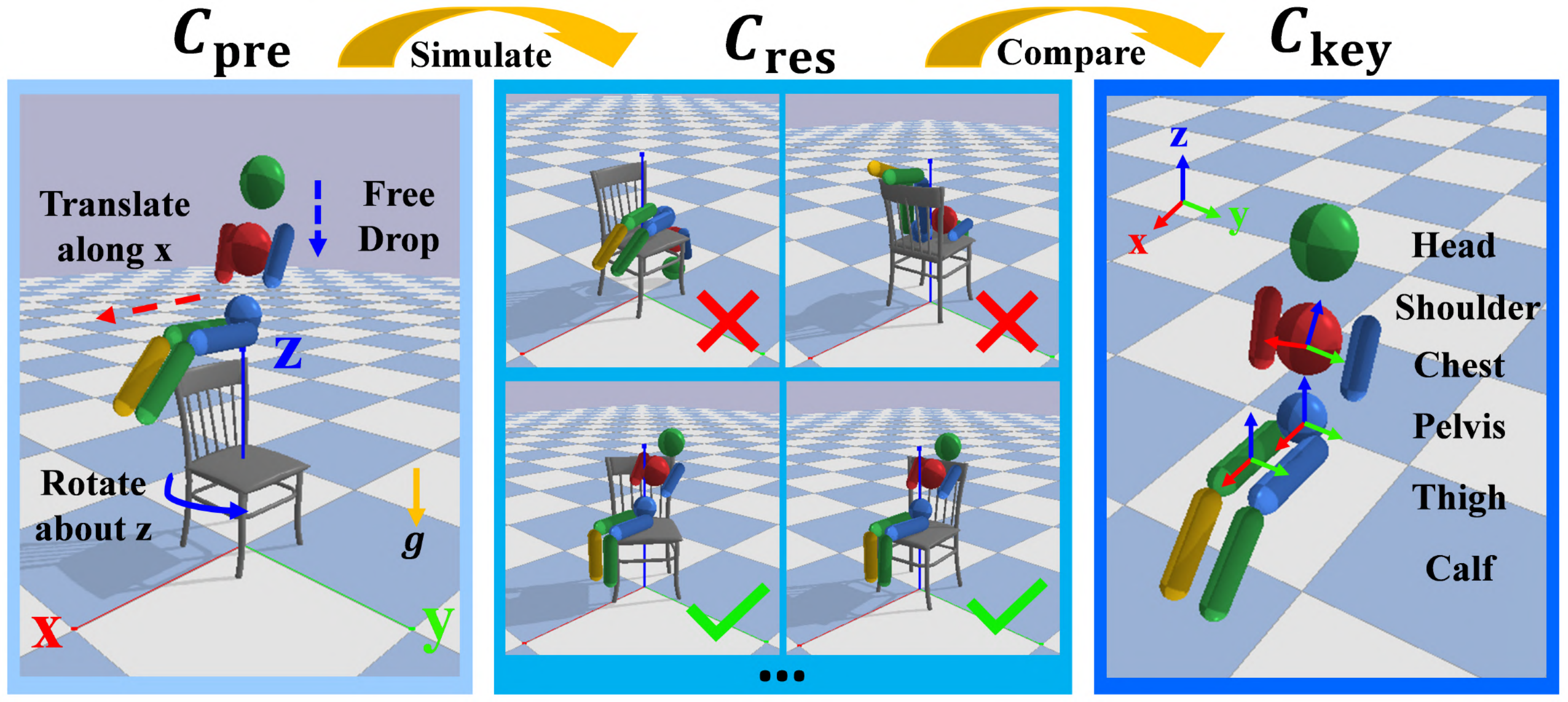}
\caption{Functional Pose Imagination. The object is placed by fixing the center of its OBB on the z-axis while keeping $z=z_s$. Before dropping, the agent is first set to a pre-sitting configuration $C_{\textrm{pre}}$ facing the x-axis as shown in the left figure. The middle figure shows four examples of $C_{\textrm{res}}$. The two with a check are regarded as \textit{correct} sittings. The agent is passively dropped onto the object. The right figure shows the key sitting configuration $C_{\textrm{key}}$. The articulated human body consists of 9 links with 18 joints. The three coordinate frames on the agent show the three link frames for computing the link rotation score $L$ (Equation (\ref{eq:6})).}
% $C_{\textrm{pre}}$ and $C_{\textrm{key}}$ are similar to a ``relaxing sitting" configuration.

\label{figure4}
\vspace{-0.2cm}
\end{figure}

The agent's resultant configuration $C_{\textrm{res}}$ in each sitting trial of a stable pose $g_s$ is compared to a key sitting configuration $C_{\textrm{key}}$ (Figure \ref{figure4}) to estimate the \textit{sitting quality} $S$ of $g_s$. The sitting affordance model estimates $S$ with four criteria: joint angle score, link rotation score, sitting height, and number of contact points.

\textbf{Joint Angle Score.}
    The joint angles of a configuration can be described with a vector $\pmb{\theta} \in \mathbb{R}^{18}$. We calculate the weighted L1 distance between the joint angle vector of $C_{\textrm{res}}$ (denoted as $\pmb{\theta}_{\textrm{res}}$) and $C_{\textrm{key}}$ (denoted as $\pmb{\theta}_{\textrm{key}}$) to obtain the joint angle score $J$ of a sitting trial:
    \begin{equation}
        J = \sum_{i} w_{J}^i |\theta_{\textrm{res}}^{i} - \theta_{\textrm{key}}^{i}|
    \end{equation}
    where $w_{J}^i$ is the weight of the i-th joint; $\theta_{\textrm{res}}^{i}$ and $\theta_{\textrm{key}}^{i}$ are the i-th element of $\pmb{\theta}_{\textrm{res}}$ and $\pmb{\theta}_{\textrm{key}}$, respectively. We assign weights based on the joint's relevance to sitting: chest-pelvis, pelvis-thigh, and thigh-calf have positive weights; other joints have zero weights. $w_{J}^i$ increases by threefold if $|\theta_{\textrm{res}}^{i} - \theta_{\textrm{key}}^{i}|$ exceeds a threshold.
    
\textbf{Link Rotation Score.}
    According to the \textit{interaction-based definition}, the chest and pelvis are more or less upright while the thigh are horizontal when the agent is sitting. Therefore, we consider the link's rotation of $C_{\textrm{res}}$ in the world frame to obtain the link rotation score $L$ of a sitting trial:
    \begin{equation}
        L = \sum_{i} w_{L}^{i} (1 - \textbf{z}_{\textrm{res}}^{i}\cdot\textbf{z}_{\textrm{key}}^{i})
    \label{eq:6}
    \end{equation}
    where $w_{L}^i$ is the weight for the i-th link; $\textbf{z}_{\textrm{key}}^{i}$ is the z-axis unit vector of the frame attached to the i-th link in $C_{\textrm{key}}$ (Figure \ref{figure4}); $\textbf{z}_{\textrm{res}}^{i}$ is the corresponding vector in $C_{\textrm{res}}$. We assign positive weights to the chest, pelvis and left/right thigh; zero weights for other links. $w_{L}^i$ increases by threefold if $(1 - \textbf{z}_{\textrm{res}}^{i} \cdot \textbf{z}_{\textrm{key}}^{i})$ exceeds a threshold.

\textbf{Sitting Height.}
    Sitting height is an important factor in the interaction-based definition of chairs. Therefore, we measure the sitting height $H$ in each sitting trial.
    
\textbf{Number of Contact Points.}
    The interaction-based definition also specifies that the buttocks and the back (corresponding to the left/right thigh and chest link) are in contact with the object when sitting. Therefore, we count the number of contact points of the agent's head $P_{H}$, chest $P_{C}$, and left/right thigh $P_{LT}$,  $P_{RT}$ in each sitting trial. 

\textbf{Detection.}    
    The $C_{\textrm{res}}$ of a sitting trial is considered as a \textit{correct} sitting (Figure \ref{figure4}) if all the followings are satisfied:
    \begin{equation*}
        J < J_{\textrm{thr}}\\, L < L_{\textrm{thr}}\\, H \in (H_{\textrm{min}}, H_{\textrm{max}})
    \end{equation*}
    \begin{equation*}
        (P_{H} + P_{C}) \cdot P_{LT} \cdot P_{RT} > 0    
    \end{equation*}
    \begin{equation*}
        P_{H} + P_{C} + P_{LT} + P_{RT} \geq P_{\textrm{thr}}
    \end{equation*}
    $J_{\textrm{thr}}$, $L_{\textrm{thr}}$, $H_{\textrm{min}}$, $H_{\textrm{max}}$, and $P_{\textrm{thr}}$ are thresholds corresponding to the four criteria. We count the number of correct sittings $N$ and calculate the mean sitting height $\xoverline{H}$ of the 54 sitting trials for each $g_s$. We select the $g_s$ with the largest value of $N\xoverline{H}$ to be the candidate functional pose $g_{\textrm{cand}}$ of the object.
The sitting quality $S_{\textrm{cand}}$ of $g_{\textrm{cand}}$ is defined as:
\begin{equation}
S_{\textrm{cand}} = \frac{N_{\textrm{cand}}\xoverline{H}_{\textrm{cand}}^{2}}{\xoverline{J}_{\textrm{cand}}\xoverline{L}_{\textrm{cand}}}
\label{eq:7}
\end{equation}
where $N_{\textrm{cand}}$ is the number of correct sittings of $g_{\textrm{cand}}$; $\xoverline{H}_{\textrm{cand}}, \xoverline{J}_{\textrm{cand}}$, and $\xoverline{L}_{\textrm{cand}}$ are the average of the sitting height, the joint angle score, and the link rotation score of all the correct sitting trials of $g_{\textrm{cand}}$.
We regard the object as a chair and $g_\textrm{cand}$ as a functional pose if:
\begin{equation*}
    ((S_{\textrm{cand}} > S_{\textrm{thr}}) \vee (N_{\textrm{cand}} \geq N_{\textrm{thr}})) \wedge (S_{\textrm{cand}} > s_{\textrm{thr}} N_{\textrm{cand}})    
\end{equation*}
where $S_{\textrm{thr}}$, $N_{\textrm{thr}}$, and $s_{\textrm{thr}}$ are three thresholds.
% Given a sitting height $H \in (H_{\textrm{min}}, H_{\textrm{max}})$, the closer $C_{\textrm{res}}$ to $C_{\textrm{key}}$, the larger the value of $S_{\textrm{cand}}$.
We experimented with different definitions of $S_{\textrm{cand}}$ with the training examples and found that using a power of $\xoverline{H}_{\textrm{cand}}^{2}$ in $S_{\textrm{cand}}$ gives better performance than linear in $\xoverline{H}_{\textrm{cand}}$.
We consider $S_{\textrm{cand}} > s_{\textrm{thr}} N_{\textrm{cand}}$ because we want to discard those objects with a low average sitting quality for each correct sitting of its $g_{\textrm{cand}}$.

\section{Evaluation}

\begin{figure}[!htp]
\centering
\includegraphics[scale = 0.27]{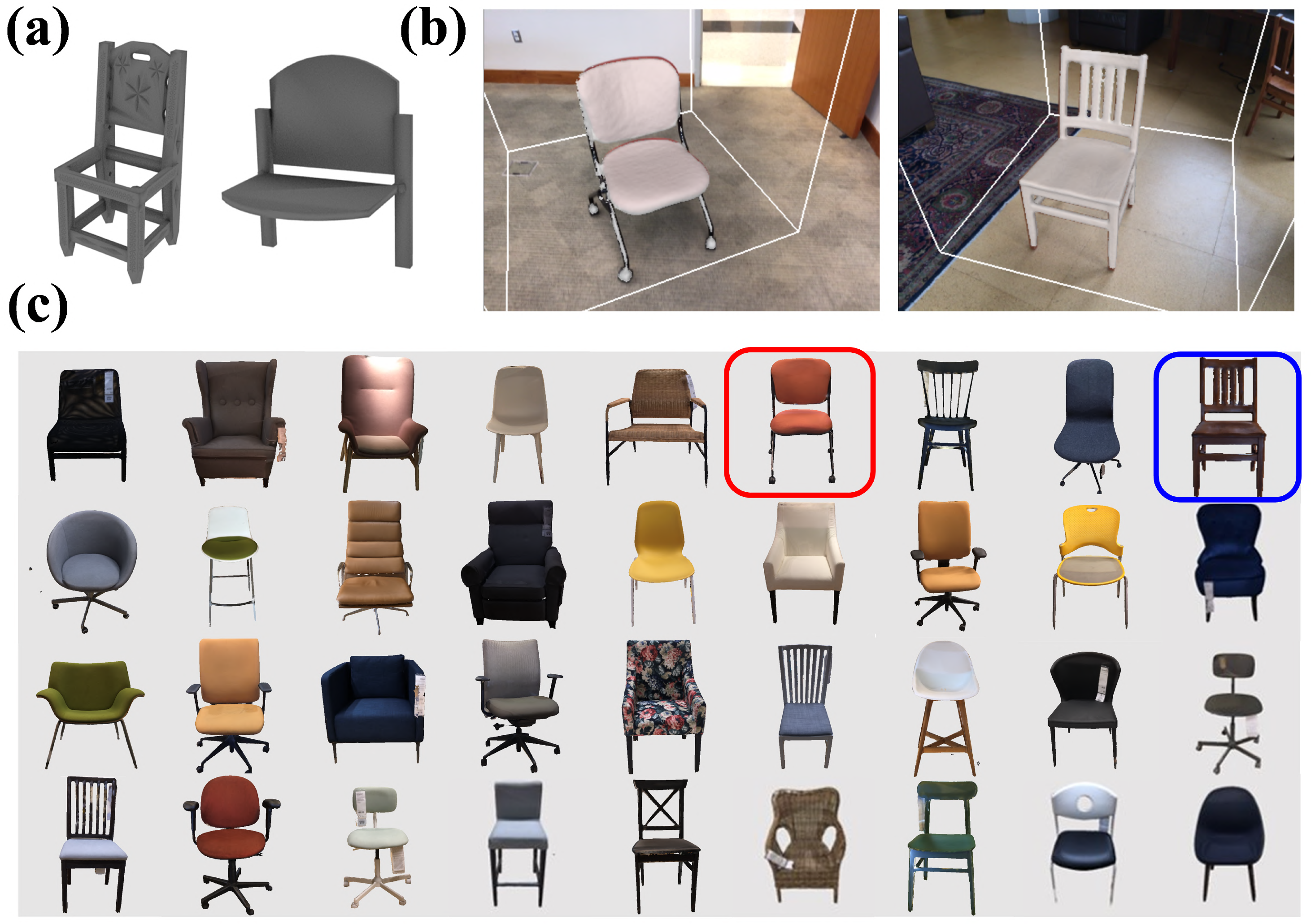}
\caption{Data. (a) Two objects from the test set (chair class) of \cite{wu20153d}. Although their appearances are very close to chairs, one does not have a seat while the other only has two legs. (b) Scanning GUI of the Occipital Structure Sensor. We use the segmentation algorithm provided by the sensor SDK to extract chairs from the scenes. The chairs in the left and right figures correspond to the ones in the red and blue boxes in (c), respectively. (c) Examples of our scanned real chair models.}
\label{figure5}
\vspace{-0.2cm}
\end{figure}

\subsection{Physics Integration}
We use PyBullet \cite{coumanspybullet} as the physical engine for simulation. The flat plane, the object, and the agent are imported with URDFs which specify the mass, center of mass, inertia matrix, friction coefficient and joint properties. We use the default Coulomb friction model. The collision between the object and the flat plane is modelled as almost inelastic (coefficient of restitution $e_{\textrm{object}} = 0.1$). The collision between the agent and the object is modelled as perfectly inelastic ($e_\textrm{human} = 0$).

\subsection{Data}
Our synthetic models are sourced from the Princeton ModelNet40 dataset \cite{wu20153d}. We use 30 objects from the chair class of the training set for training. Our synthetic test data consist of 300 synthetic models extracted from the test set: 100 objects from the chair class and 200 objects from 20 non-chair object classes (10 objects per class). The non-chair objects are common household objects, \textit{e.g.}, TV stands, beds, desks, tables, bathtubs, and cups. Since the scale of the object is essential in determining its affordances \cite{gibson2014ecological}, we scale all the synthetic objects to an appropriate size of their corresponding classes. Our real test data consist of 50 real chairs scanned with the Occipital Structure Sensor (Figure \ref{figure5}(b)(c)). They are not scaled. As we are interested in finding the functional pose from an arbitrary initial orientation, we further randomly orient all the objects in the synthetic and real test data.

Since both synthetic and real chair models have no annotations of \textit{functional pose}, we recruited 10 volunteers to annotate the functional pose via the PyBullet GUI interface. For each chair model, the volunteer first determined whether it is sittable. A flat ground was then presented and the volunteer was asked, ``How would you place it on the ground to sit on it?" The annotated pose was recorded as the functional pose annotation.
It is worth mentioning that although the 100 chair models in the synthetic test data are labeled as chair in \cite{wu20153d}, two of them are labelled as not sittable by the human annotator (Figure \ref{figure5}(a)). In our experiment, we consider them as non-chairs because they cannot afford sitting.

\subsection{Training}
The OBB transformation is computed using the Trimesh Python Library \cite{trimesh}. Volumetric Hierarchical Approximate Convex Decomposition (V-HACD) \cite{mamou2016volumetric} is applied to decompose the transformed object into a set of convex hulls for collision detection in the simulation. We use MeshLab to compute the mass, center of mass, and inertia matrix of the object assuming a uniform density of 600 kg/m$^3$ (comparable to the density of wood).
We manually drop the agent onto the 30 training examples and obtain $C_{\textrm{pre}}$, $C_{\textrm{key}}$, $\Delta R_{\textrm{thr}}^{\textrm{es}}$, $\Delta p_{\textrm{thr}}^{\textrm{es}}$, $\Delta R_{\textrm{thr}}^{\textrm{s}}$, $\Delta p_{\textrm{thr}}^{\textrm{s}}$, $\Delta o_{\textrm{thr}}$, $J_{\textrm{thr}}$, $L_{\textrm{thr}}$, $H_{\textrm{min}}$, $H_{\textrm{max}}$, and $P_{\textrm{thr}}$ which maximize the \textit{sitting quality} $S$ and number of correct sittings $N$ for all the training examples. In detection, we set $S_{\textrm{thr}}=5\textrm{m}^2/\textrm{rad}$, $N_{\textrm{thr}}=4$, and $s_{\textrm{thr}}=0.1\textrm{m}^2/\textrm{rad}$ according to the value of $S$ and $N$ of the training examples.

\section{Results}
We implement our method with Python. We perform the evaluation on a computer running Intel Core i7-8700 @ 3.2GHz CPU. Our single-threaded unoptimized implementation takes about 1.5 minutes to analyze a single raw model. The imagination accounts for about 70 seconds of that time.
 
\subsection{Chair Classification}
We first perform the chair vs non-chair classification on the synthetic test data. We compare our method with two state-of-the-art appearance-based 3D object classifiers \cite{kanezaki2018rotationnet, su2018deeper}. Both methods leverage deep neural networks to classify a 3D object with multiple views of the object. \cite{kanezaki2018rotationnet} uses 20 views encompassing the object without the upright orientation assumption. \cite{su2018deeper} uses 12 views and assumes that the object is oriented upright. We note that in the original training settings of \cite{kanezaki2018rotationnet} and \cite{su2018deeper}, validation (model selection) is performed on the test set of \cite{wu20153d}. Since our synthetic test data are extracted from the test set of \cite{wu20153d}, for fair comparison, we retrain the two baselines with 80\% of the original training set and use the rest 20\% as the validation set for model selection. We use the retrained models for evaluation. The training set of \cite{wu20153d} contains 9843 models of 40 classes in total (the chair class has 889 models). We use the rendered images and the rendering code provided by the authors for training and rendering our test data for evaluation, respectively.

\
\begin{table}[!htp]
\vspace{-0.6cm}
\caption{Chair Classification Accuracy On Synthetic Data(\%)}
\centering
{
\begin{tabular}{c c  c }
\toprule
 Method & Object Orientation & Accuracy \\
 \arrayrulecolor{black}\midrule
 Kanezaki \textit{et al.} \cite{kanezaki2018rotationnet} & upright  & 99.3 \\
 Su \textit{et al.}\cite{su2018deeper} & upright     & 99.0 \\ 
 \arrayrulecolor{black!30}\midrule
 Kanezaki \textit{et al.} \cite{kanezaki2018rotationnet} & random  & 83.0 \\
  Su \textit{et al.} \cite{su2018deeper} & random   & 86.3 \\ 
 Ours & random                          & \textbf{97.0}\\
\arrayrulecolor{black}\bottomrule
\end{tabular}
}
\label{table:1}
\vspace{-0.1cm}
\end{table}

We use the overall classification accuracy (Table \ref{table:1}) and mean average precision (Figure \ref{figure6}) for quantitative evaluation. We test \cite{kanezaki2018rotationnet} and \cite{su2018deeper} with and without the upright orientation assumption. Although we only use 30 synthetic chair models for training, our algorithm achieves the highest classification accuracy when the upright orientation is \textit{not assumed}. 
% \cite{kanezaki2018rotationnet} outperforms us by only 2.3\% when \textit{assuming an upright orientation} for the test data. 
Note that in Table \ref{table:1}, the performance of both \cite{kanezaki2018rotationnet} and \cite{su2018deeper} drops dramatically when the upright orientation assumption is removed.

\subsection{Functional Pose Prediction}
We also evaluate the \textit{functional pose} prediction of our method on both synthetic and real test data. The rotation matrix $R$ of a rigid body transformation $g=(R, \textbf{p})$ can be specified as
$R=[\textbf{v}^{x}, \textbf{v}^{y}, \textbf{v}^{z}]^{T}$,
where $\textbf{v}^{x}$, $\textbf{v}^{y}$, $\textbf{v}^{z}$ are the unit vectors of the world frame's three axes in the body frame. Since a chair in the functional pose would be \textit{equivalently stable} regardless of its $\textbf{v}^{x}$, $\textbf{v}^{y}$, $x$, and $y$, we compare the $\textbf{v}^z$ and $z$ of the predicted functional pose (denoted as $\textbf{v}_{\textrm{pred}}^{{z}}$ and $z_{\textrm{pred}}$) to those of the functional pose annotation (denoted as $\textbf{v}_{\textrm{ann}}^z$ and $z_{\textrm{ann}}$). If $1-\textbf{v}_{\textrm{pred}}^{{z}}\textbf{v}_{\textrm{ann}}^{{z}} \leq 0.01$ and $|z_{\textrm{pred}} - z_{\textrm{ann}}| \leq 0.01\textrm{m}$, we consider the prediction correct and incorrect if otherwise. If a non-chair is classified as a chair, we count it as a false positive. If a chair is classified as a non-chair or its functional pose prediction is incorrect, we count it as a false negative. The precision and recall of the functional pose prediction on the synthetic data are both 94.9\%. For the 50 real chair models, the recall of the functional pose prediction is 100\%. We can see that our method generalizes well to the real data.

\subsection{Rotation Metric Sensitivity}
In Sec. \ref{sec:stable_pose}, we use the Frobenius norm to compute the distance between two rotations (Equation (\ref{eq:1}) and (\ref{eq:3})). 
To probe our method's sensitivity against rotation metrics, we replace the Frobenius norm with a geometric-based metric  \cite{chirikjian2016harmonic} which defines the distance between two rotations to be the angle of rotation from one to the other, \textit{i.e.}, $d(R_1, R_2) = |\theta_{12}|$. The angle of rotation $\theta_{12}$ is derived from:
\begin{equation*}
    e^{\widehat{w}_{12}\theta_{12}} = R_1^{-1}R_2
\end{equation*}
\begin{equation*}
    \theta_{12} = \|(\textrm{log}(R_1^{-1}R_2))^\vee\|
\end{equation*}
where $\widehat{w}_{12} \in so(3)$ (or $w_{12} \in \mathbb{R}^3$) specifies the axis of rotation between $R_1$ and $R_2$ \cite{chirikjian2016harmonic}. The geometric structure of $SO(3)$ can be identified as an upper hemisphere of a unit sphere in $\mathbb{R}^{4}$. The Frobenius norm measures the straight-line distance between two points on the sphere while the geometric-based norm measures the arcs length. The chair classification accuracy and the precision and recall of the functional pose prediction on the corresponding test data are the same as those using the Frobenius norm. Therefore, our method is not sensitive to the choice of rotation metrics.

\begin{figure}[!htp]
    \centering
    \includegraphics[scale=0.279]{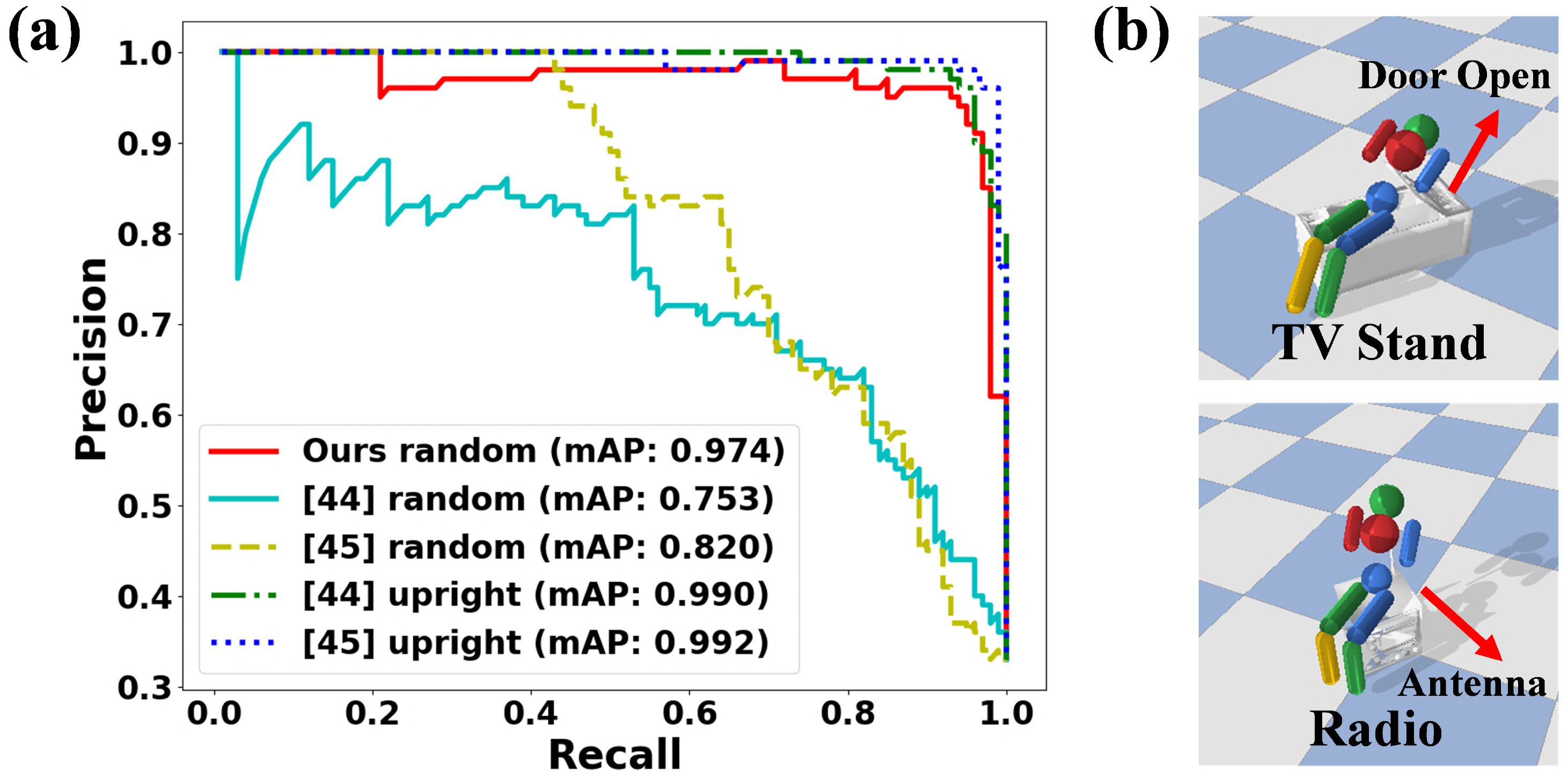}
    \caption{Results. (a) Chair classification precision-recall curve. ``random" and ``upright" refer to the test object orientation. In our case, the sitting quality $S$ is used to calculate the mean average precision (mAP). Our method outperforms the deep learning methods when the upright orientation is not assumed. (b) Two examples of false positive in the chair classification owing to the single rigid body modeling. The agent's back rests on the opened closet door in the TV stand case and on the antenna in the radio case.}
    \label{figure6}
\vspace{-0.3cm}
\end{figure}

\section{Discussion and Future Work}
 In contrast to the black box nature of deep learning methods, our method is fully explainable. It is able to \textit{explain} the sitting affordance of an object by inferring the \textit{functional pose} in addition to classification. Indeed, these two tasks are closely related in terms of object perception. Our method captures the most essential cue of chairs, the sitting affordance, with only 30 training data while the compared deep learning methods use thousands of examples to distill appearance cues which may not be adaptable between different data domains (\textit{e.g.}, random vs upright). The appearance-based methods are also limited when the object appears similar to a class but fails to afford the functionality of the class. \cite{kanezaki2018rotationnet} and \cite{su2018deeper} classify the two objects in Figure \ref{figure5}(a) as chairs when tested with the upright orientation assumption. Our method classifies both as non-chairs because no \textit{functional pose} can be found for each.

Although we have incorporated many physical properties, all the test models in the simulations are considered as single rigid bodies (Figure \ref{figure6}(b)). Future studies can incorporate more detailed physical properties (\textit{e.g.}, joints and elasticity) of the object of interest in the simulation and explore more object classes with complex object affordances. Also, the dropping simulations with different initial orientations in the stable pose imagination are independent to each other. This is also true for the sitting simulations in the functional pose imagination. Future studies can parallelize these two imagination processes via multi-threading to reduce processing time.

\section{Conclusion}
In this paper, we propose a novel method for robots to \textit{imagine} an object's affordance using physical simulations. The class of chair is chosen here to illustrate a more general paradigm of interaction-based object affordance reasoning. We simulate dropping the object to find the stable poses and the sitting interaction between a simulated human agent and the object to find the \textit{functional pose} which affords the functionality of sitting. The object and the agent are both subject to physical and geometric constraints. The \textit{imagination} of object affordances is used as a cue for chair classification and functional pose prediction. In chair classification, results show that our method outperforms two state-of-the-art deep learning methods on the synthetic test data when the upright orientation of the object is not assumed. The functional pose predictions of our method on both synthetic and real data align well with human judgements. We hope that our method will serve as an effective approach to guide robot-object interaction in future research.

\section*{Acknowledgment}
This work was performed under Office of Naval Research Award N00014-17-1-2142 and National Science Foundation grant IIS-1619050.

\bibliographystyle{IEEEtran}
\bibliography{reference.bib}
\end{document}